# QSM-RimDS: A detection and segmentation tool for paramagnetic rim lesions in multiple sclerosis

Ha Luu[1,6], PhD, Mert Sisman[1,2], MS, Ilhami Kovanlikaya[1], MD, Tam Vu[3], Pascal Spincemaille[1], PhD, Francesca Bagnato[4], MD PhD, Yi Wang[1], PhD, Susan A. Gauthier[5], DO, Thanh D. Nguyen[1], PhD

[1]*Department of Radiology, Weill Cornell Medicine, New York, NY, USA*

[2]*Department of Electrical and Computer Engineering, Cornell University, Ithaca, NY, USA*

[3]*Yale University, New Haven, CT, USA*

[4]*Department of Neurology, Vanderbilt University Medical Center, Vanderbilt, TN, USA*

[5]*Department of Neurology, Weill Cornell Medicine, New York, NY, USA*

[6]*University of Engineering and Technology, VNU, Hanoi, Vietnam*

## Abstract

**Background**: Paramagnetic rim lesions (PRLs) are an emerging biomarker in multiple sclerosis (MS). Manual identification and rim segmentation of PRLs on quantitative susceptibility mapping (QSM) images are time-consuming. Deep learning-based QSM-RimNet can provide automated PRL detection, but this method does not provide rim segmentation for microglial density quantification and requires precise QSM lesion masks.

**Purpose**: To develop a U-Net-based QSM-RimDS method for joint PRL detection and rim segmentation using readily available T2-weighted (T2W) fluid-attenuated inversion recovery (FLAIR) lesion masks.

**Study Type**: Cross-sectional retrospective.

**Population**: 256 MS patients (177 women [69.1%], 79 men [30.9%]; mean age, 46.2 ± 11.8 years, age range, 19-78 years).

**Field Strength/Sequence**: 3 Tesla/3D T1-weighted, 3D T2W FLAIR, and 3D multi-echo gradient echo sequences.

**Assessment**: Two expert readers performed PRL classification and rim segmentation as the reference. Dice similarity coefficient (DSC) was used to assess the agreement between rim segmentation obtained by QSM-RimDS and the manual expert segmentation. The PRL detection performances of QSM-RimDS and QSM-RimNet were evaluated using receiver operating characteristic (ROC) and precision-recall (PR) plots in a five-fold cross validation.

**Statistical Tests**: None

**Results**: A total of 260 PRLs (3.3%) and 7720 non-PRLs (96.7%) were identified by the readers. Compared to the expert rim segmentation, QSM-RimDS provided a mean DSC of 0.57 ± 0.02 with moderate to high agreement (DSC ≥ 0.5) in 73.8 ± 5.7% of PRLs over five folds. QSM-RimDS produced better and more consistent detection performance with a mean area under curve (AUC) of 0.754 ± 0.037 vs. 0.514 ± 0.121 by QSM-RimNet (46.7% improvement) on PR plots, and 0.956 ± 0.034 vs. 0.908 ± 0.073 (5.3% improvement) on ROC plots.

**Data Conclusion**: QSM-RimDS improves PRL detection accuracy compared to QSM-RimNet and unlike QSM-RimNet can provide reasonably accurate rim segmentation.

**Keywords**: multiple sclerosis, paramagnetic rim lesion (PRL), quantitative susceptibility mapping (QSM), segmentation, QSM-RimNet, U-Net

## INTRODUCTION

Multiple sclerosis (MS) is a chronic autoimmune disease of the central nervous system and a leading cause of neurologic disability in young adults (1). Paramagnetic rim lesions (PRLs) form a subset of chronic active MS brain lesions with persistent smoldering immune activity (2) and are associated with progressive tissue damage and clinical disability (3-5). PRLs are characterized histologically by a relatively inactive core and a dense layer of iron-laden activated microglia and macrophages at the lesion periphery (6,7). These lesions can be identified in vivo as having a rim appearance on iron-sensitive gradient echo MR images obtained by high-pass filtered phase imaging (8-11) or quantitative susceptibility mapping (QSM) (12-14). Compared to phase imaging, QSM can provide more accurate PRL detection (15) and direct quantification of iron-laden activated microglia (7) which can be used as a diagnostic or treatment biomarker for innate immune response in MS.

Currently, identification and rim segmentation of PRLs on QSM images by human readers is time-consuming and prone to reader bias (5,16). A deep learning and radiomics-based PRL detection method called QSM-RimNet (17) has been developed to automate rim identification. However, this algorithm requires precisely traced lesion masks on QSM, which are difficult to obtain, to achieve the best performance. Furthermore, QSM-RimNet cannot provide rim segmentation for activated microglia quantification. Here, we propose a U-Net-based method called QSM-RimDS to perform joint PRL rim detection and segmentation on QSM images using readily available T2-weighted (T2W) fluid-attenuated inversion recovery (FLAIR) lesion masks. We evaluated PRL detection performance of QSM-RimDS compared with QSM-RimNet in MS patients using classification by human experts as the reference.

## MATERIALS AND METHODS

This was a cross-sectional retrospective study conducted in 256 MS patients (177 women [69.1%], 79 men [30.9%]; mean age, 46.2 ± 11.8 years, age range, 19-78 years) selected from an ongoing MS imaging and clinical research database. The database was created following approval of the local institutional review board and written informed consent was obtained from all participants.

**MRI acquisition and post-processing**

Subjects were imaged on 3T scanners (Siemens Healthineers, Erlangen, Germany) using a product 20-channel head/neck coil. The brain MRI protocol contained 3D T1-weighted (T1W) sequence for anatomical definition, 3D T2W FLAIR sequence for lesion identification, and 3D multi-echo gradient echo (mGRE) for PRL detection using QSM. The following imaging parameters were used: 1) sagittal T1W: TR/TE/TI = 2300.0/2.3/900 ms, flip angle (FA) = 8°, bandwidth = 200 Hz/pixel, voxel size = 1.0 mm isotropic, parallel imaging factor (R) = 2.0, scan time = 5:21 min; 2) sagittal T2W FLAIR: TR/TE/TI = 7600/448/2450 ms, FA = 90°, bandwidth = 781 Hz/pixel, ΔTE = 3.42 ms, turbo factor = 284, voxel size = 1.0 mm isotropic, R = 4.0, scan time = 5:21 min; 3) axial mGRE: TE1/ΔTE = 6.28/4.06 ms, number of TEs = 8, TR = 40 ms, FA = 15°, bandwidth = 260 Hz/pixel, voxel size = 0.4x0.4x1 $mm^3$, R = 2.0, scan time = 5:05 min.

QSM images were reconstructed from the complex mGRE data using an automated morphology-enabled dipole inversion with uniform global cerebrospinal fluid zero reference (MEDI+0) algorithm (18). FLAIR lesion masks were automatically created using ALL-Net lesion segmentation tool (19), manually corrected if necessary by an expert reader, and aligned to the echo-combined mGRE magnitude image, which exists in in the image space as the QSM image, using ANTs image registration toolbox (20).

**Expert PRL classification**

Two expert readers independently reviewed FLAIR-defined lesions on QSM images. Following recent 2024 consensus guidelines (21), a lesion was classified as PRL if it had a hyperintense rim surrounding approximately two-thirds or more of the lesion perimeter on at least two consecutive slices. For network training, lesions were labeled as PRL if both readers agreed, otherwise they were labeled as non-PRL. Subsequently, the rim areas of each lesion labeled as PRL were manually traced and jointly confirmed by the same readers.

**QSM-RimDS training and evaluation**

The proposed QSM-RimDS framework consists of a deep learning-based convolutional neural network trained to segment a hyperintense lesion rim followed by a threshold-based binary classifier for PRL detection (Fig.1). We adopted a nested cross-validation (80%/20% split) approach (22) which consists of an inner loop for network training and an outer loop for tuning the thresholds for PRL classification.

For rim segmentation, we used 3D nnU-Net, a state-of-the-art network segmentation tool with automatic configuration of preprocessing, network architecture, and training (23). The network input is a QSM image patch (64x64x24 voxels) centered on the lesion and the output is a probabilistic map which is thresholded by a tunable parameter $\lambda_p$ to generate a rim segmentation mask. Within each patch, a 3 mm-dilated FLAIR lesion mask was applied to exclude image voxels that lie well outside the lesion. To handle the class imbalance between PRLs and non-PRLs, the set of PRLs was augmented such that the number of augmented PRLs including the original PRLs was equivalent to that of non-PRLs in the training set. Network training was performed using the default five-fold cross-validation (200 epochs per fold) using a weighted sum of Dice similarity coefficient (DSC) and Binary Cross Entropy as the loss function (the optimal weight was determined empirically to maximize the mean DSC over one fold). Network prediction was obtained by averaging the probability output of the models trained over five folds. DSC was used to assess the agreement between rim segmentation obtained by QSM-RimDS and the manual expert segmentation.

For PRL detection, the length of the segmented rim was measured by applying morphological thinning operation to the rim segmentation mask, from which a rim ratio index (rim length/FLAIR lesion perimeter) was calculated on a per slice basis. A lesion was identified as PRL if the rim ratio was greater than a tunable threshold $\lambda_r$ on at least two consecutive slices. Given our goal of developing a highly sensitive detection tool for PRL screening, the two thresholding parameters ($\lambda_p$ and $\lambda_r$) were tuned using grid search to maximize the detection precision for a minimum detection sensitivity of 90% in a five-fold cross-validation. For comparison, QSM-RimNet was trained and validated on the same QSM patches and FLAIR lesion masks using the same nested five-fold cross-validation approach. The PRL detection performances of QSM-RimDS and QSM-RimNet were evaluated using receiver operating characteristics (ROC) and precision-recall (PR) plots. Since PRLs are relatively rare, the PR plot is generally regarded as a better measure of detection performance for practical applications.

## RESULTS

A total of 260 PRLs (3.3%) and 7720 non-PRLs (96.7%) were identified by the two readers from 256 subjects, of whom 92 (35.9%) had at least one PRL. Among non-PRLs, 177 lesions (2.3%) were classified as PRL by only one reader. There was a substantial agreement between the two readers with regards to PRL identification (Cohen's $\kappa = 0.73$).

Compared to manual rim segmentation, QSM-RimDS achieved a mean DSC of 0.57 ± 0.02 and moderate to high agreement (DSC ≥ 0.5) in 73.8 ± 5.7% of PRLs over five folds. Figure 2 shows an example of rim segmentation of a PRL correctly identified by both QSM-RimDS and QSM-RimNet, demonstrating accurate delineation of the hyperintense rim area by nnU-Net. The segmentation obtained by QSM-RimDS was in good agreement with the expert segmentation (DSC = 0.66) despite the rim's proximity to structures with similar susceptibility values such as the cortical ribbon and central veins (Fig.2). Figure 3 shows an example of a false positive PRL detected by QSM-RimDS and QSM-RimNet. This lesion had a thin rim that was segmented well by QSM-RimDS; however, the two readers disagreed on the status of this lesion most likely due to the weak QSM contrast between the lesion rim and core. On average, non-PRLs that were initially identified as PRL by one reader accounted for 54.4 ± 9.0% and 17.0 ± 2.5% of false positive predictions by QSM-RimDS and QSM-RimNet, respectively.

When tuned to achieve ≥90% detection sensitivity for PRL screening purposes, QSM-RimNet yielded a mean detection sensitivity of 88.8 ± 4.6%, but did not reach the desired 90% sensitivity in one out of the five folds and had a low mean precision of 26.3 ± 3.5%. In contrast, QSM-RimDS was able to achieve a high detection sensitivity of 90.4% over all five folds with mean detection precision of 63.5 ± 3.1% (141% improvement) (Table 1). Figure 4 shows two lesion examples which illustrate the improvement in PRL detection obtained by the proposed QSM-RimDS compared to QSM-RimNet. In the first example (top row), a PRL with a prominent central vein and partial rim appearance was missed by QSM-RimNet (false negative) but correctly identified by QSM-RimDS (true positive). In the second example (bottom row), a non-PRL was incorrectly classified as PRL by QSM-RimNet (false positive) while QSM-RimDS provided a correct detection (true negative). Overall, QSM-RimDS produced better and more consistent detection performance with a mean AUC of 0.754 ± 0.037 vs. 0.514 ± 0.121 by QSM-RimNet (46.7% improvement) on PR plots (Fig.5) and 0.956 ± 0.034 vs. 0.908 ± 0.073 (5.3% improvement) on ROC plots (Fig.6).

## DISCUSSION

Our results show that the proposed QSM-RimDS tool based on the nnU-Net design provides more accurate and robust PRL detection than QSM-RimNet. QSM-RimDS offers two important benefits previously lacking in QSM-RimNet: eliminating the need for precise lesion masks on QSM and providing rim segmentation for QSM quantification of iron-rich microglia as a biomarker of innate immune activity. Various efficient machine learning algorithms have been developed for FLAIR lesion mask extraction (17,24). With PRLs gaining broader acceptance as a diagnostic and treatment biomarker in MS, most notably reflected in the 2024 McDonald criteria revision, an automated PRL detection and rim segmentation tool such as QSM-RimDS has great potential for alleviating the burden of manual labor on human readers and fostering seamless clinical translation.

In this study, our goal was to develop a PRL screening tool with high detection sensitivity (≥90%) and moderate precision (50-70%) which can substantially reduce the number of lesions that need to be reviewed by human readers. This is possible because PRLs are quite rare and account for only about 5% of all lesions. For example, assuming QSM-RimDS sensitivity of 90% and precision of 63%, the number of PRL candidates that need to be reviewed following QSM-RimDS prediction is reduced by more than 20-fold from the original total of 7980 lesions in our study to only 371 lesions. Future studies are needed to evaluate the time saved by using QSM-RimDS as a computer-aided PRL screening tool in routine radiological workflow.

## Limitations

Our study has several limitations. First, the lesion patch size of 64x64x24 voxels may be too small for large confluent FLAIR lesions. Possible solutions include increasing the patch size for training or breaking such a lesion into smaller overlapping patches. Second, a lesion was designated as PRL only if both readers agreed, which helped to ensure that only PRLs with clear rim definition were selected for network training. A better design would be to include more expert readers to increase PRL selection confidence, which will likely improve network training and reduce the number of false positives. Finally, our study was conducted using a relatively small sample size from a single center. Further studies are warranted to evaluate the generalizability of our network model on data acquired from multiple centers and scanner platforms.

## Conclusion

QSM-RimDS improves PRL detection accuracy compared to QSM-RimNet and unlike QSM-RimNet can provide reasonably accurate rim segmentation for microglial density quantification.

**Table 1**. Comparison of PRL detection performances provided by QSM-RimNet and the proposed QSM-RimDS operating as a PRL screening tool with a target sensitivity of at least 90%. The mean and standard deviation of performance metrics measured over five folds are shown.

| | **Sensitivity** | **Specificity** | **Precision** | **Accuracy** |
|---|---|---|---|---|
| QSM-RimNet | 88.8 ± 4.6% | 90.8 ± 3.2% | 26.3 ± 6.5% | 90.8 ± 3.2% |
| QSM-RimDS | 90.4 ± 0.0% | 98.2 ± 0.2% | 63.5 ± 3.1% | 98.0 ± 0.2% |

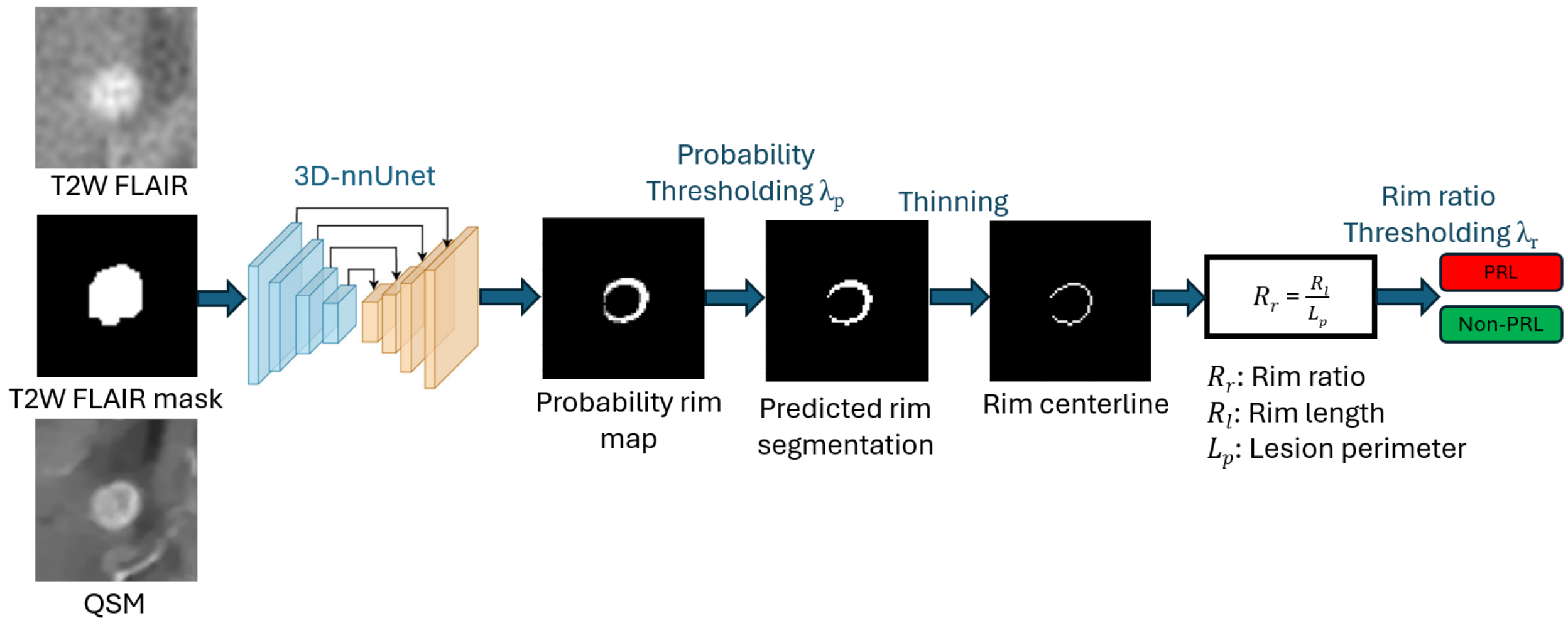

**Figure 1**. The proposed QSM-RimDS framework for joint PRL segmentation and detection on QSM. The thresholds for network output rim probability map ($\lambda_p$) and rim ratio ($\lambda_r$) are tunable parameters and in this work were selected to maximize the detection precision for a target detection sensitivity of at least 90%.

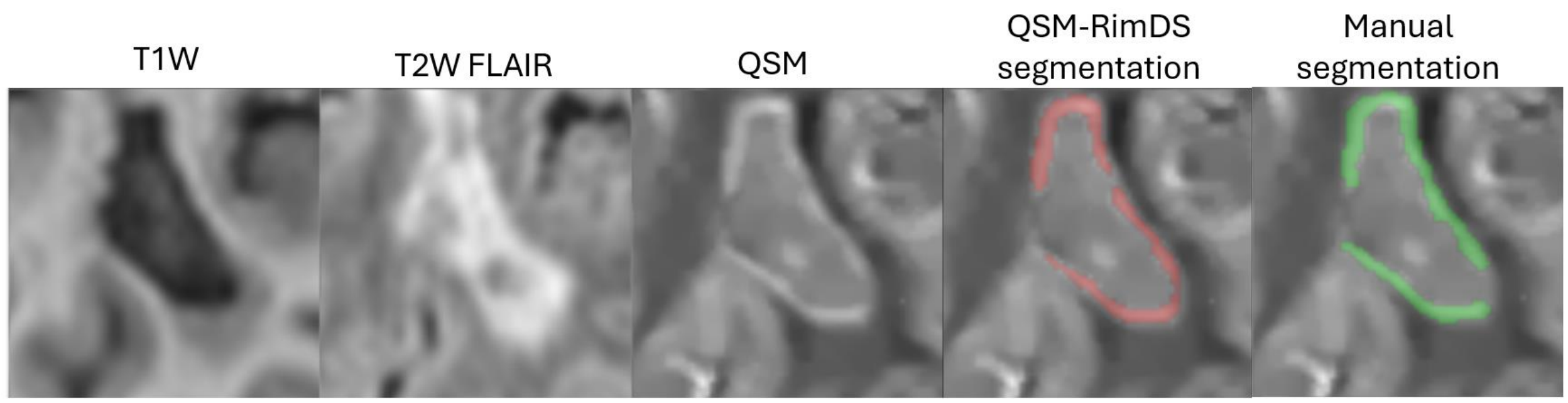

**Figure 2**. An example of axial T1W, T2W FLAIR and QSM images of a juxtacortical MS lesion with paramagnetic rim appearance. The rim segmentation produced by the proposed QSM-RimDS method (red) and manual segmentation by experts (green) are overlaid on the QSM image, showing good agreement with a DSC of 0.66.

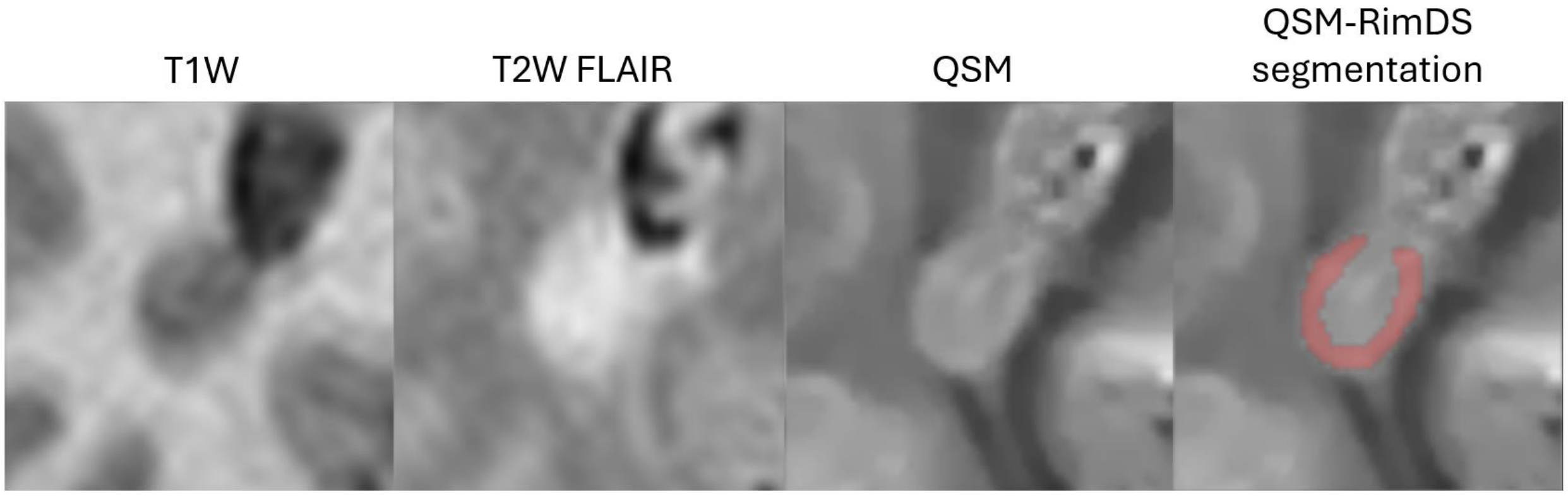

**Figure 3**. An example of a non-PRL with a thin rim appearance and weak contrast between lesion rim and core, which was detected as PRL (false positive) by QSM-RimNet and QSM-RimDS. The rim segmentation produced by QSM-RimDS is shown in red. Note that one reader identified this lesion as PRL.

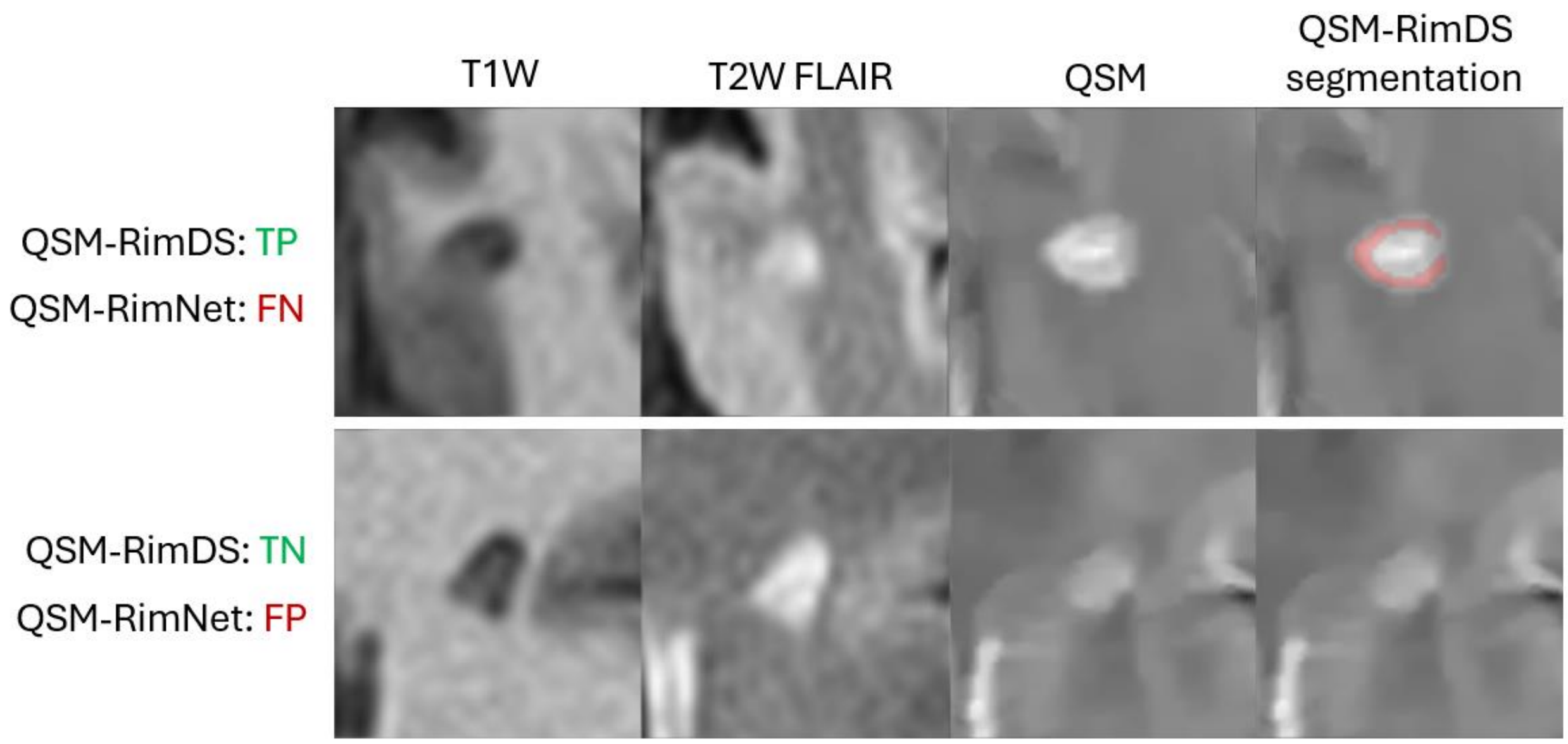

**Figure 4**. Example of a PRL (top row) and a non-PRL (bottom row) lesion that were incorrectly classified by QSM-RimNet as non-PRL (false negative) and PRL (false positive), respectively. By contrast, the proposed QSM-RimDS correctly predicted the PRL status of these lesions.

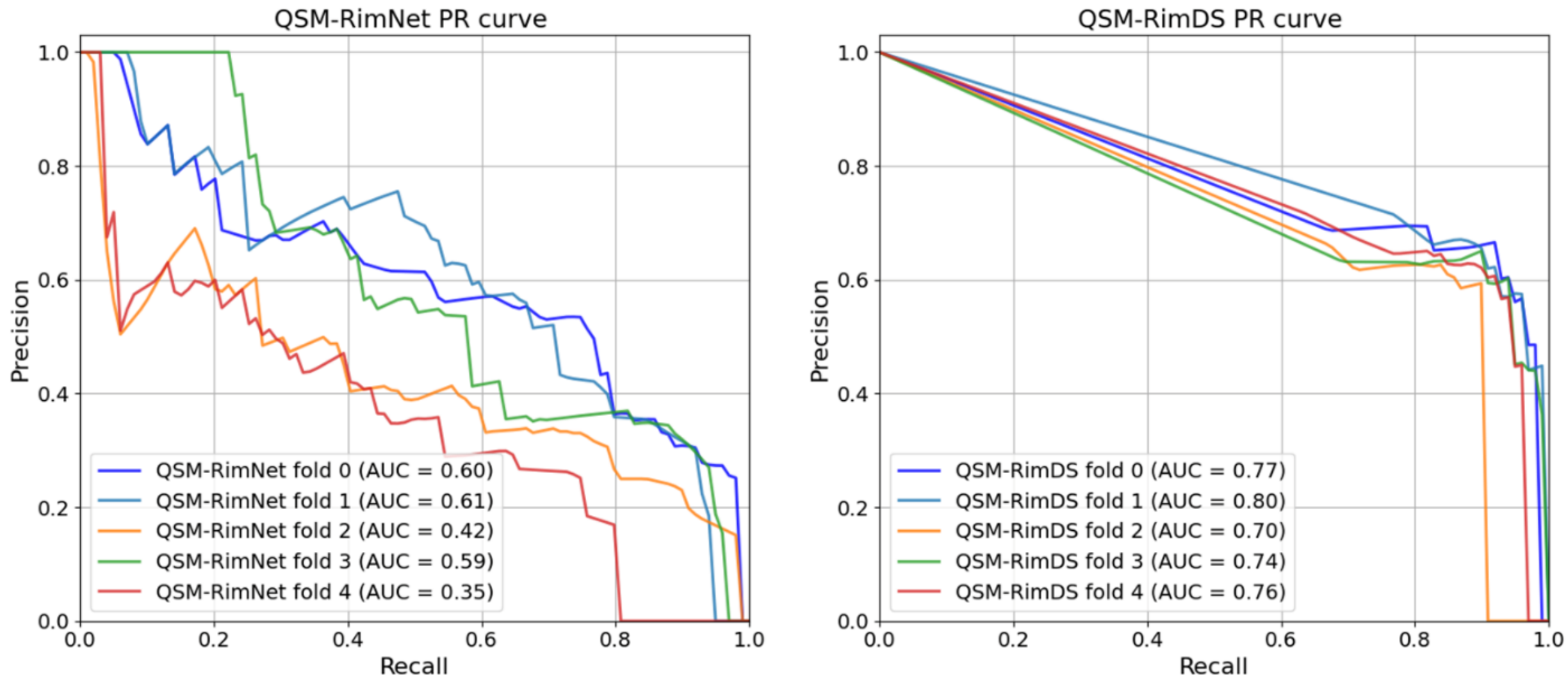

**Figure 5**. Precision-recall (PR) plots of QSM-RimNet vs. QSM-RimDS obtained for PRL detection over five folds.

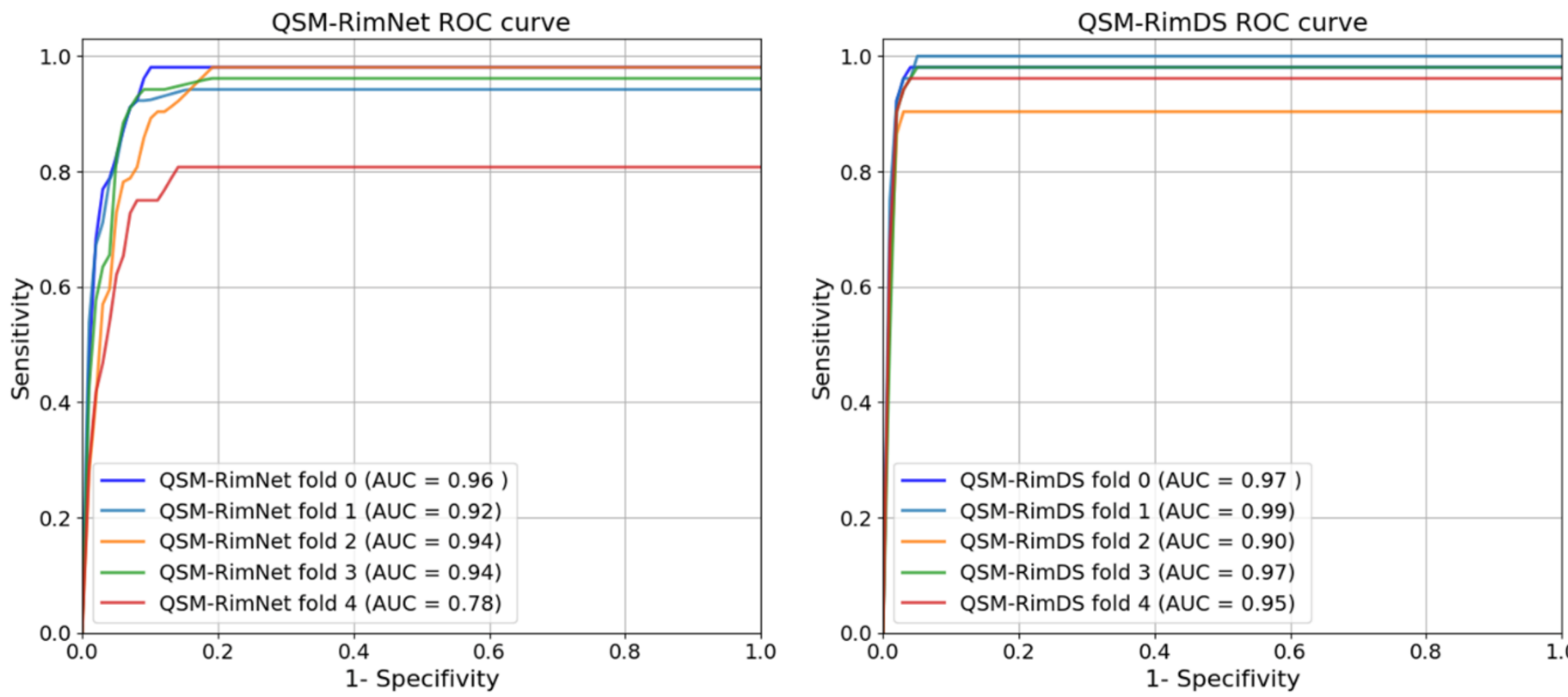

**Figure 6**. Receiver operating characteristics (ROC) plots of QSM-RimNet vs. QSM-RimDS obtained for PRL detection over five folds.